\documentclass[9pt,conference]{IEEEtran}
\usepackage{amsmath,amssymb,amsfonts,amsthm,enumitem}

\usepackage{amssymb}
\usepackage{graphicx}
\usepackage{textcomp}
\usepackage{float}
\usepackage{placeins}
\usepackage{multirow}
\usepackage{dblfloatfix}
\usepackage{pifont}
\usepackage{colortbl}
\usepackage[normalem]{ulem}
\useunder{\uline}{\ul}{}
\usepackage{booktabs}
\usepackage{balance}
\usepackage{comment}
\usepackage{subcaption}
\usepackage{hyphenat}
\usepackage{amsmath}
\usepackage{tablefootnote}
\usepackage{algorithm}
\usepackage{algpseudocode}
\usepackage{siunitx}
\usepackage{url}

\algrenewcommand\algorithmiccomment[1]{\texttt{// #1}}

\AtBeginDocument{%
  \providecommand\BibTeX{{%
    \normalfont B\kern-0.5em{\scshape i\kern-0.25em b}\kern-0.8em\TeX}}}

\begin{document}

\title{RIFT: A Scalable Methodology for LLM Accelerator Fault Assessment using Reinforcement Learning}
\author{\IEEEauthorblockN{Khurram Khalil, Muhammad Mahad Khaliq, Khaza Anuarul Hoque}
\IEEEauthorblockA{\textit{Department of Electrical Engineering and Computer Science}\\ \textit{University of Missouri-Columbia, USA}\\
\{khurram.khalil, mkfqm, hoquek\}@missouri.edu}
}  
\maketitle


\begin{abstract}

The massive scale of modern AI accelerators presents critical challenges to traditional fault assessment methodologies, which face prohibitive computational costs and provide poor coverage of critical failure modes. This paper introduces RIFT (Reinforcement Learning-guided Intelligent Fault Targeting), a scalable framework that automates the discovery of minimal, high-impact fault scenarios for efficient design-time fault assessment. RIFT transforms the complex search for worst-case faults into a sequential decision-making problem, combining hybrid sensitivity analysis for search space pruning with reinforcement learning to intelligently generate minimal, high-impact test suites. Evaluated on billion-parameter Large Language Model (LLM) workloads using NVIDIA A100 GPUs, RIFT achieves a \textbf{2.2$\times$} fault assessment speedup over evolutionary methods and reduces the required test vector volume by over \textbf{99\%} compared to random fault injection, all while achieving \textbf{superior fault coverage}. The proposed framework also provides actionable data to enable intelligent hardware protection strategies, demonstrating that RIFT-guided selective error correction code provides a \textbf{12.8$\times$} improvement in \textbf{cost-effectiveness} (coverage per unit area) compared to uniform triple modular redundancy protection. RIFT automatically generates UVM-compliant verification artifacts, ensuring its findings are directly actionable and integrable into commercial RTL verification workflows.
\end{abstract}


\section{Introduction}
\label{sec:introduction}

The recent advent of Large Language Models (LLMs) with hundreds of billions of parameters has had a transformative impact on computing, but has also introduced unprecedented computational demands~\cite{sze2017efficient}. The sheer scale of these models makes their execution on traditional processors infeasible, necessitating the development of specialized \emph{AI accelerators}. These massively parallel architectures, such as GPUs and TPUs, feature vast on-chip memory systems and complex dataflows designed to handle the immense throughput requirements of modern AI workloads~\cite{jouppi2017datacenter}. While such specialized AI accelerators are powerful, ensuring their reliability against \emph{hardware faults} presents a formidable challenge for the Electronic Design Automation (EDA) industry. To put this challenge into perspective, consider a representative large foundation model. A model with $m$ billion parameters, quantized to $n$-bit integers, presents over $m \times n \times 10^9$ potential single-bit fault locations. However, the true complexity lies in combinatorial faults, as worst-case failures often arise from a few synergistic bit-flips. The number of possible $k$-bit fault combinations in such a model is given by the binomial coefficient $\binom{m \cdot n \cdot 10^9}{k}$. For a typical 8-bit quantized, 8-billion parameter model ($m=8, n=8$) and a minimal attack vector of just five bits ($k=5$), this value exceeds $\binom{6.4 \times 10^{10}}{5} \approx 10^{50}$ possible combinations—a search space so vast that exhaustive fault assessment is computationally intractable~\cite{clarke2018model, rebaudengo2005fault}.

\begin{figure*}[ht]
    \centering
    \includegraphics[width=0.8\textwidth]{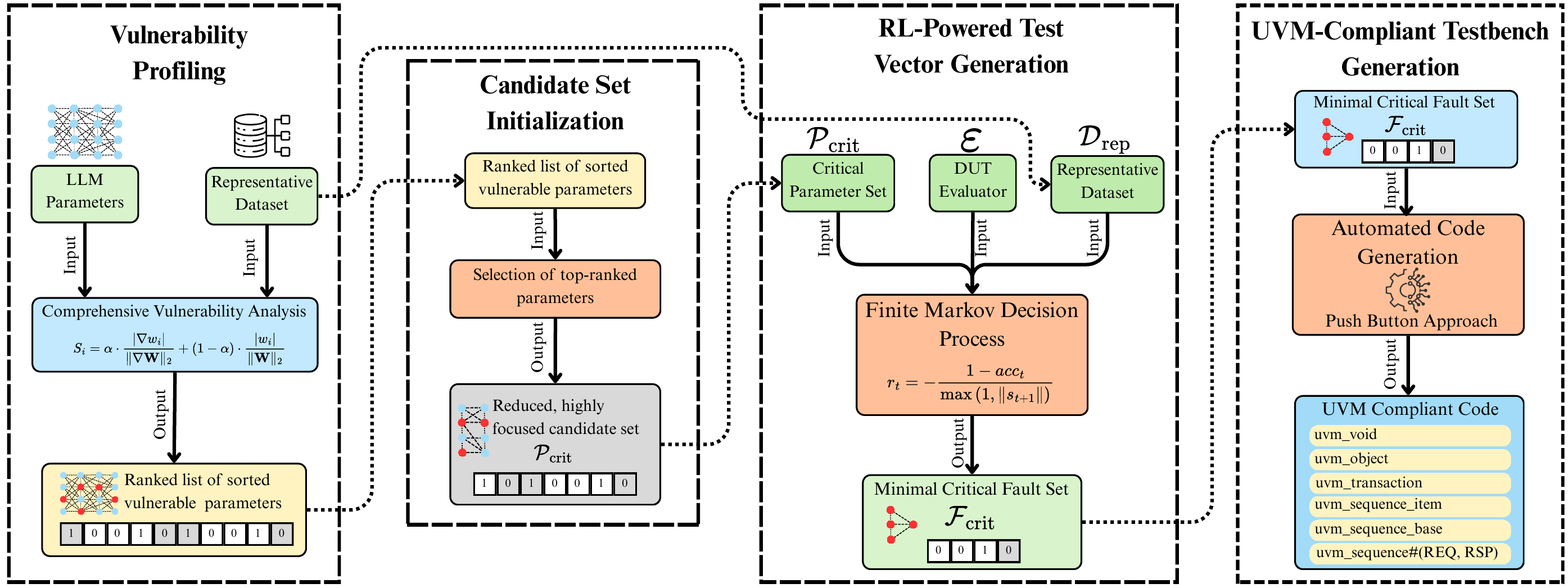}
    \caption{Overview of the RIFT framework.}
    \label{fig:framework}
\end{figure*}

The industry-standard for addressing this challenge has long relied on Random Fault Injection (RFI) and its variations, which provide a practical but statistically inefficient means of exploring the fault space~\cite{hsiao1998fault, mukherjee2008architecture}. RFI requires prohibitive simulation time to achieve even modest coverage and is fundamentally ill-suited to uncovering the sparse, worst-case failure modes that often pose the greatest risk, a challenge compounded by the difficulty in estimating coverage and confidence for such methods~\cite{pattabiraman2008coverage}. At the other end of the spectrum, formal methods---powered by advances in Boolean satisfiability (SAT) solving~\cite{moskewicz2001chaff} and deployed in techniques like bounded model checking~\cite{clarke2001bounded} and symbolic test generation~\cite{geist1996coverage}---offer mathematical rigor but are computationally intractable for the billion-transistor scale of modern AI accelerators due to state-space explosion~\cite{clarke2018model}. 
Machine learning has also been successfully applied to the related domain of \emph{functional verification} to generate test stimuli and find design bugs~\cite{wu2024survey, gadde2024efficient}. these approaches are fundamentally stimulus generators for logic validation, not fault campaign generators for reliability assessment.
Within the specific domain of AI accelerators, research has evolved beyond these general-purpose techniques to characterize the unique vulnerabilities of Deep Neural Networks (DNNs), a field extensively surveyed in recent literature~\cite{ahmadilivani2024slr}.
Foundational frameworks like Ares~\cite{reagen2018ares} and DeepHammer~\cite{yao2020deephammer} established that reliability analysis could be significantly improved by identifying architectural ``hotspots" using static or dynamic analysis of parameter importance. Subsequent work has confirmed that the consequences of faults are highly non-uniform and can propagate extensively through DNN layers~\cite{li2017understanding}. However, these methods were designed for and validated on DNNs with millions of parameters; their computational cost does not readily scale to the billions of parameters in modern LLMs.
To tackle the combinatorial search space in LLMs, the most relevant work has emerged from the security domain, with methods like the evolutionary algorithm GenBFA~\cite{das2024attentionbreaker} and the heuristic-based PrisonBreak~\cite{coalson2024prisonbreak} targeting minimal bit-flips for adversarial attacks. 
These approaches, while more effective than random search, lack the adaptive, sequential learning capabilities needed to efficiently navigate the complex, non-linear fault space of modern AI hardware. Thus, a fundamental challenge remains: \emph{how to efficiently and systematically search the vast combinatorial space of bit locations within billion-parameter LLMs to identify the minimal, high-impact bit sets capable of inducing catastrophic failure?}

This paper introduces \emph{Reinforcement Learning-guided Intelligent Fault Targeting (RIFT)}, a novel framework that transforms the computationally expensive search for critical faults into an efficient sequential decision-making problem solved via RL. RIFT enables intelligent and automated exploration of the vast fault space, moving beyond brute-force and random sampling to efficiently identify the minimal set of high-impact faults needed for a comprehensive fault assessment, while being architecture-agnostic. For instance, RIFT identifies the minimal, high-impact bit sets capable of inducing catastrophic failure in LLMs \textbf{2.2$\times$} faster than the state-of-the-art evolutionary method~\cite{das2024attentionbreaker} while achieving \textbf{superior fault coverage} and \textbf{$>$99\%} test vector reduction, enabling practical fault assessment of LLM models with billion-parameters. Identifying such potential fault locations enables RIFT to also provide actionable data to enable intelligent hardware protection strategies. For instance, we show that applying selective protection to the RIFT-identified fault locations is up to \textbf{12.8$\times$} \textbf{cost-effective}  when compared to the traditional uniform triple modular redundancy (TMR) scheme. Furthermore, the practical CAD integration of RIFT is demonstrated through the automated generation of \textbf{UVM-compliant testbenches} which can be directly used with commercial verification environments with minimal effort.


\section{Problem Formulation}
\label{sec:problem_formulation}

To formalize the fault assessment challenge, we define the following notations. Let $\mathcal{M}$ denote the LLM under analysis and $\mathbf{W} = \{w_1, w_2, \dots, w_n\}$ be its set of learnable parameters. The set $\mathbf{W}$ encompasses all parameter groups within the LLM targeted for faults, including the self-attention projection matrices (Query, Key, Value, and Output), the feed-forward networks, and the normalization layers. 
The set of all addressable bit locations within these parameters constitutes the fault space, $\mathcal{B}(\mathbf{W})$. A fault set is a subset $\mathcal{F} \subseteq \mathcal{B}(\mathbf{W})$, representing the specific bit-flips to be injected. Applying this fault set to the original model yields a perturbed model, $\mathcal{M}_{\mathcal{F}}$. The model's functional correctness is measured by an evaluation metric, $\text{acc}(\mathcal{M}, \mathcal{D}_{rep})$, on a representative dataset, $\mathcal{D}_{rep}$, and catastrophic failure is defined by a performance threshold, $\tau$.

\textbf{Given:}
An LLM $\mathcal{M}$ with its parameter set $\mathbf{W}$, a representative evaluation dataset $\mathcal{D}_{rep}$, an evaluation metric $\text{acc}(\cdot)$, and catastrophic failure threshold $\tau$.

\textbf{Find:}
A minimal, high-impact bit set $\mathcal{F}_{crit}$ capable of inducing catastrophic failure. The optimization problem is formally stated as:
\begin{equation}
\begin{aligned}
\mathcal{F}_{crit} = \min_{\mathcal{F} \subseteq \mathcal{B}(\mathbf{W})} \quad & |\mathcal{F}| \\
\textrm{s.t.} \quad & \text{acc}(\mathcal{M}_{\mathcal{F}}, \mathcal{D}_{rep}) \le \tau
\end{aligned}
\label{eq:optimization_problem}
\end{equation}
Solving this problem is computationally intractable due to the combinatorial explosion of the fault space $\mathcal{B}(\mathbf{W})$. The RIFT framework is therefore designed to find an effective and computationally feasible approximate solution.

\section{The Proposed RIFT Framework}
\label{sec:methodology}
The proposed RIFT framework, conceptually illustrated in Figure~\ref{fig:framework}, provides a structured and scalable approach for automating the discovery of minimal, high-impact fault scenarios in AI accelerators executing LLMs, formulated as the optimization problem defined in Section~\ref{sec:problem_formulation}.
The framework functions as an algorithmic funnel that systematically narrows the vast fault space of the Design Under Test (DUT), where the fault sites correspond to the bit locations of the target LLM's parameters $\mathbf{W}$, to generate a compact test suite.

The process begins with \emph{Phase 1, Vulnerability Profiling}, where a hybrid metric combining static parameter importance and dynamic functional sensitivity is used to create a comprehensive, ranked profile of all learnable LLM parameters. The dynamic sensitivity component is computed by evaluating parameter gradients with respect to the model's loss on a small but representative dataset $\mathcal{D}_{rep}$. The input to this phase is the complete set of the LLM's parameters $\mathbf{W}$ and $\mathcal{D}_{rep}$, and its output is a ranked list of all parameters according to their sensitivity scores. Next, in \emph{Phase 2, Candidate Set Initialization}, the framework leverages this profile to select a high-sensitivity subset of parameters from the most sensitive regions, further narrowing the search space. The input to this phase is the ranked parameter list from Phase 1, and its output is a reduced, high-potential candidate set of parameter indices to be targeted. Finally, \emph{Phase 3, RL-Powered Test Vector Generation}, formulates the search for the optimal fault combination as a sequential decision-making problem. Taking the candidate set from Phase 2 as its input, an RL agent refines this set via reward-guided exploration, which serves as the final output of the framework. The following subsections detail each phase of the RIFT framework.

\subsection{Phase 1: Vulnerability Profiling}
\label{subsec:phase1_profiling}

The initial phase of RIFT performs a comprehensive vulnerability analysis of the DUT to create a quantitative fault susceptibility profile. 
This analysis is guided by our target fault model, which consists of targeted, single bit-flips occurring in memory-stored parameter values. We adopt the bit-flip model as it is a fundamental representation of data corruption in memory systems and serves as a foundational abstraction for the effects of various physical hardware failures~\cite{reagen2018ares,li2017understanding}.

For each parameter $w_i$ in the DUT, we compute a hybrid sensitivity score, $S_i$, that combines its static importance (magnitude) and its dynamic importance (gradient information):
\begin{equation}
S_i = \alpha \cdot \frac{|\nabla w_i|}{||\nabla \mathbf{W}||_2} + (1 - \alpha) \cdot \frac{|w_i|}{||\mathbf{W}||_2},
\label{eq:sensitivity}
\end{equation}

where $\alpha$ is a mixing coefficient, $\nabla w_i$ is the gradient of the functional loss with respect to parameter $w_i$, and normalization is performed across the L2-norm of the entire parameter vector $\mathbf{W}$ and its corresponding gradient vector $\nabla \mathbf{W}$. Concurrently, we perform Memory Hierarchy Mapping by analyzing simulation traces or hardware performance counters to correlate logical parameters with their physical memory locations. This identifies memory access hotspots, and parameters in high-traffic regions receive additional weighting in their vulnerability score. The output of this phase is a single, ranked list of all parameters in the DUT, sorted by their composite vulnerability, effectively pruning the design space by providing a quantitative map of fault susceptibility.

\subsection{Phase 2: Candidate Set Initialization}
\label{subsec:phase2_reduction}

The ranked list of all LLM parameters, sorted by composite vulnerability from Phase 1, serves as the input to this second phase. The primary goal of this phase is to drastically reduce the search space by isolating the most sensitive parameters for the subsequent, computationally intensive RL exploration.
This is achieved by selecting a small fraction of the top-ranked parameters, a process governed by a selection rate hyperparameter, $\rho$. The value of $\rho$ represents a trade-off, balancing the need for a computationally tractable search space against ensuring the candidate set is comprehensive enough to contain the critical failure modes.

This selection process consistently localizes the most sensitive parameters to specific, computationally critical architectural components known to be computationally critical in AI accelerators.
The output of this phase is a highly focused candidate set, $\mathcal{P}_{crit}$, which contains the indices of these top-ranked parameters. This focused set makes the final RL exploration phase computationally tractable, even for billion-parameter models.

\begin{algorithm}[t]
\caption{RIFT Algorithm for Minimal Test Vector Generation}
\footnotesize
\label{alg:rift_algorithm}
\begin{algorithmic}[1]
\State \textbf{Input:} Critical parameter set $\mathcal{P}_{crit}$, DUT evaluator $\mathcal{E}$, representative dataset $\mathcal{D}_{rep}$, max episodes $E_{max}$, max episode length $T_{max}$, learning rate $\alpha_{rl}$, discount factor $\gamma$, exploration rate $\epsilon$
\State \textbf{Output:} Minimal critical fault set (test suite) $\mathcal{F}_{crit}$
\State Initialize Q-table $Q(s, a)$
\State Initialize best reward $r_{best} \leftarrow -\infty$, $\mathcal{F}_{crit} \leftarrow \emptyset$
\For{episode = 1 to $E_{max}$}
    \State Initialize state $s_0 \leftarrow \emptyset$ \Comment{Start with empty fault list}
    \For{t = 0 to $T_{max}-1$}
        \State Define action space $A_t = \{add(p) : p \in \mathcal{P}_{crit} \setminus s_t\} \cup \{remove(p) : p \in s_t\}$
        \State Select action $a_t \in A_t$ using $\epsilon$-greedy policy from $Q(\phi(s_t), \cdot)$
        \State \Comment{$\phi(s_t)$ creates a canonical, hashable representation of state $s_t$}
        \State Apply action $a_t$ to generate next state $s_{t+1}$
        \State Evaluate DUT: $\text{acc}_t \leftarrow \mathcal{E}(s_{t+1}, \mathcal{D}_{rep})$
        \State Calculate reward $r_t$ using Equation~\ref{eq:reward}
        \If{$r_t < r_{best}$} \Comment{Found a more impactful and/or smaller test suite}
            \State $r_{best} \leftarrow r_t$, $\mathcal{F}_{crit} \leftarrow s_{t+1}$
        \EndIf
        \State Update Q-table using the Bellman equation:
        \State $Q(\phi(s_t), a_t) \leftarrow Q(\phi(s_t), a_t) + \alpha_{rl}[r_t + \gamma \max_{a'} Q(\phi(s_{t+1}), a') - Q(\phi(s_t), a_t)]$
        \State $s_t \leftarrow s_{t+1}$
    \EndFor
\EndFor
\State \textbf{return} $\mathcal{F}_{crit}$
\end{algorithmic}
\end{algorithm}
\subsection{Phase 3: RL-Powered Test Vector Generation}
\label{subsec:phase3_rl}

The final phase employs an RL agent to solve the optimization problem defined in Equation~\ref{eq:optimization_problem}. We formalize this search as a finite Markov Decision Process (MDP), which is detailed in Algorithm~\ref{alg:rift_algorithm}. The algorithm takes as input the candidate parameter set $\mathcal{P}_{crit}$, a DUT evaluator $\mathcal{E}$, the representative dataset $\mathcal{D}_{rep}$, and several RL hyperparameters. First, the algorithm initializes a Q-table to store state-action values, along with variables to track the best-found fault set $\mathcal{F}_{crit}$ and its corresponding reward, $r_{best}$, (Lines 3-4). The agent's training proceeds over a maximum of $E_{max}$ training episodes (Line 5), where each episode consists of a sequence of actions over a series of up to $T_{max}$ steps, indexed by $t$.

At the beginning of each episode, the agent's state, $s_t$, which represents the current fault list, is initialized to an empty set (line 6). At each step $t$ (line 7), the agent chooses from a set of available actions, $A_t$, which consists of either adding a new parameter from $\mathcal{P}_{crit}$ to the current fault list or removing an existing one (line 8). The action $a_t$ is selected using an $\epsilon$-greedy policy, where $\epsilon$ is the exploration rate, based on the current Q-table (line 9). Applying action $a_t$ transitions the agent from state $s_t$ to a new state $s_{t+1}$ (line 11). This new fault set is then evaluated on the DUT to determine the resulting functional accuracy, $\text{acc}_t$ (line 12). This accuracy is used to calculate a reward, denoted as $r_t$:
\begin{equation}
r_t = -\frac{1 - \text{acc}_t}{\max(1, |s_{t+1}|)}
\label{eq:reward}
\end{equation}
where $|s_{t+1}|$ is the number of faults in the resulting set. The agent tracks the best fault set found so far, updating $\mathcal{F}_{crit}$ whenever a new state yields a reward $r_t$ more optimal than the current best, $r_{best}$ (lines 14-16). Finally, the agent updates its Q-table using the Bellman equation, governed by the learning rate $\alpha_{rl}$ and discount factor $\gamma$, (lines 17-18) to improve its policy for future decisions. Upon completion of all episodes, the algorithm returns the final minimal, high-impact test suite $\mathcal{F}_{crit}$ (line 22), which is then formatted into industry-standard artifacts like UVM-compliant testbenches.

\subsection{UVM-Compliant Testbench Generation}

The final output of the RIFT algorithm, the minimal critical fault set $\mathcal{F}_{crit}$, serves as the input to an automated code generation stage. This stage utilizes a template-based script that parses a structured data file (e.g., a JSON file) representing the list of parameter indices and bit locations in $\mathcal{F}_{crit}$ and programmatically generates a UVM-compliant \emph{test sequence}. The resulting SystemVerilog sequence encapsulates the entire fault injection campaign within a dedicated data structure, specifically a queue (\texttt{\$[]}) of \texttt{fault\_item} objects. Each \texttt{fault\_item} object, a custom class extending \texttt{uvm\_object}, explicitly defines a target by its parameter index and the bit position to be flipped. When this sequence is run in a standard UVM \emph{testbench}, it uses the UVM \emph{configuration database} (\texttt{config\_db}) to pass this fault list to a fault injection agent. This agent, a reusable fault injection component, then drives the DUT's interfaces to inject the precise faults at the correct time during the simulated workload. This code generation flow bridges the critical gap between high-level algorithmic analysis and practical RTL fault assessment, enabling a ``push-button" approach to running targeted, high-impact fault campaigns. To validate this capability, we successfully executed RIFT-generated test sequences on a target accelerator design within the Xilinx Vivado Design Suite, confirming that our framework produces fault assessment artifacts ready for direct use in commercial CAD workflows.

\subsection{Algorithmic Complexity and Scalability}
The computational complexity of the RIFT algorithm is dominated by the RL phase. The time complexity is $O(|\mathcal{P}_{crit}| \cdot E_{max} \cdot T_{max} \cdot C_{eval})$, where $T_{max}$ is the maximum episode length and $C_{eval}$ is the cost of a single DUT evaluation. Since the size of the critical set, $|\mathcal{P}_{crit}|$, is much smaller than the total number of parameters, RIFT remains tractable. The space complexity is determined by the size of the Q-table. For practical purposes, our tabular implementation is effective for $|\mathcal{P}_{crit}|$ up to several thousand parameters, a range sufficient for covering the fault-sensitive hotspots in the tested billion-parameter models.


\begin{table}[t]
    \centering
    \caption{Fault Assessment Efficiency Comparison.
    Cov = Coverage, TV = Test Vectors, Eff = Efficiency measured as Coverage per hour (Mean $\pm$ SD over 15 runs), and SU = Speed up wr.t. RFI.}
    \label{tab:efficiency_comparison}
    \scriptsize
    \begin{tabular*}{\columnwidth}{@{\extracolsep{\fill}}lccccc@{}}
        \toprule
        \textbf{Methods} & \textbf{Cov} & \textbf{Time} & \textbf{TV} & \textbf{Eff} & \textbf{SU} \\
        & \textbf{(\%)} & \textbf{(hrs)} & & \textbf{(Cov/hr)} & \\
        \midrule
        RFI & 65.3 $\pm$ 4.1 & 1000 $\pm$ 52 & $1.2 \times 10^5$ & 0.065 & 1.0× \\
        Magnitude Ranking & 73.8 $\pm$ 3.7 & 245 $\pm$ 18 & $8.4 \times 10^3$ & 0.301 & 4.6× \\
        Gradient Selection & 79.2 $\pm$ 2.9 & 198 $\pm$ 15 & $6.1 \times 10^3$ & 0.400 & 6.2× \\
        GenBFA~\cite{das2024attentionbreaker} & 84.6 $\pm$ 3.2 & 388 $\pm$ 24 & $4.7 \times 10^3$ & 0.218 & 3.4× \\
        \midrule
        \textbf{RIFT} & \textbf{91.7 $\pm$ 2.1} & \textbf{187 $\pm$ 12} & \textbf{847 $\pm$ 73} & \textbf{0.490} & \textbf{7.5×} \\
        \bottomrule
    \end{tabular*}
\vspace{-5mm}    
\end{table}

\section{Experimental Evaluation}
\label{sec:experimental_evaluation}
This section presents a comprehensive experimental validation of RIFT framework with respect to three fundamental questions: \\
(1) \textbf{RQ1: Fault Assessment Efficiency--} Does RIFT significantly reduce the computational cost and time required for fault assessment? \\
(2) \textbf{RQ2: Reliability-Aware Design Insights--} Does RIFT provide actionable data to enable intelligent hardware protection strategies? \\
(3) \textbf{RQ3: Framework Consistency and Scalability--} Is the RIFT algorithm robust, scalable, and suitable for integration into commercial fault assessment workflows?

\subsection{Evaluation Methodology and Setup}
\label{subsec:experimental_setup}

\textbf{Simulation Platform, DUT, and Target Models:} Our evaluation utilizes a high-performance compute cluster with NVIDIA A100 80GB GPUs. The DUT is a simulated AI accelerator executing three representative LLMs as target workloads: GPT-2 Large~\cite{radford2019language}, LLaMA 3.1 8B~\cite{llama2024}, and the Mixture-of-Experts model DeepSeek-V2 7B~\cite{deepseek2024}. All models utilize 8-bit integer quantization to reflect common deployment practices~\cite{dettmers2022gpt3}.
We use the Massive Multisk Language Understanding (MMLU) benchmark~\cite{hendrycks2020measuring} and its more challenging MMLU-Pro variant~\cite{wang2024mmlupro} to assess the functional correctness of the DUT. These benchmarks measure knowledge and reasoning across 57 subjects with over 15,000 multiple-choice questions, and 14 subjects with 12,000 questions, respectively.
To represent a worst-case scenario for numerical deviation, faults are injected by flipping the most significant bit (MSB) of selected parameters during inference, which aligns with established practices in hardware reliability studies~\cite{das2024attentionbreaker, mukherjee2008architecture, nazari2024forget}.

\textbf{Baseline Methodologies for Comparison:} We compare RIFT against a comprehensive suite of four established and state-of-the-art methodologies to ensure a thorough evaluation. These baselines include: (1) widely-adopted \emph{Random Fault Injection (RFI)~\cite{mukherjee2008architecture, hsiao1998fault}}; (2) a \emph{Magnitude-Based Ranking} heuristic, representing static analysis methods similar to those used in recent work like PrisonBreak~\cite{coalson2024prisonbreak}; (3) a \emph{Gradient-Based Selection} heuristic, which implements the core principles of seminal dynamic analysis techniques like DeepHammer~\cite{li2019deephammer}; and (4) the state-of-the-art \emph{GenBFA} evolutionary search algorithm~\cite{das2024attentionbreaker}. 
To evaluate RIFT's effectiveness in guiding design space exploration, we analyze five distinct hardware protection schemes using representative overhead estimates derived from industry synthesis studies~\cite{micron2023ecc}. The analyzed schemes include a baseline with \emph{No Protection}, low-cost \emph{Parity Detection}, industry-standard \emph{ECC SECDED}, high-resilience \emph{ECC ChipKill}, and maximum-reliability \emph{Triple Modular Redundancy (TMR)}.

All baselines operate under identical computational budgets and evaluation criteria to ensure a fair comparison.

\textbf{Coverage Definition and Statistical Methodology:} Fault coverage is defined as the percentage of critical fault scenarios (those causing $>90\%$ accuracy degradation) successfully identified by each methodology within a fixed computational budget of 1,000 CPU hours. This budget was chosen to represent a practical, albeit extensive, fault assessment time window for a single analysis block in an industrial setting, providing a fair basis for comparing the efficiency of different methodologies. To provide a standardized and hardware-agnostic measure of computational cost, all runtimes are reported in CPU hours, reflecting the total resource allocation for each experiments. To ensure statistical rigor, all experiments are repeated 15 times with different random seeds, and significance is assessed via Welch's t-test with Bonferroni correction for multiple comparisons.


\subsection{RQ1: Fault Assessment Efficiency}
\label{subsec:assessment_efficiency}

This section quantifies and analyzes the RIFT's impact on fault assessment productivity by measuring the fault discovery efficiency against established baselines. For this, in Table~\ref{tab:efficiency_comparison}, we report several key metrics, including fault coverage achieved (Cov), the computational time required, the number of test vectors generated (TV), and the overall efficiency (Eff) for each compared method.
The data in Table~\ref{tab:efficiency_comparison} shows that RIFT outperforms all baseline methods across every key metric. For instance, let us consider the Efficiency metric (Coverage per hour), which captures the trade-off between coverage quality and computational cost. RIFT achieves an efficiency score of 0.490, which is \textbf{7.5$\times$ higher than RFI} (0.065) and \textbf{2.2$\times$ higher than the state-of-the-art GenBFA evolutionary search} (0.218). This demonstrates a significant improvement in the ability to find critical faults within a fixed computational budget.

A direct comparison of the time and test vectors required to achieve high coverage provides a more granular view of this efficiency gain. While the RFI baseline required 1000 CPU hours to achieve a modest 65.3\% coverage, RIFT reached a far superior \textbf{91.7\% coverage in only 187 CPU hours}.
Furthermore, RIFT achieves this with an exceptionally compact test suite, generating an average of only 847 fault scenarios. This stands in stark contrast to the thousands of test vectors required by heuristic methods, including GenBFA (ranging from 4,700 to 8,400) and the over 100,000 vectors needed for the RFI campaign, representing a greater than \textbf{99\% reduction in test vector volume} against the widely adapted industry standard.

This superior performance stems from RIFT's algorithmic design. Unlike the undirected exploration of RFI or the purely heuristic guidance of other baselines, RIFT's RL agent intelligently learns and adapts its search strategy. This allows it to systematically discard unproductive regions of the vast fault space and converge efficiently on the sparse, high-impact fault combinations that constitute the greatest risk, thereby delivering a transformative improvement in fault assessment productivity.

\begin{table}[b]
    \centering
    \caption{Critical Vulnerability Discovery Across Diverse DUTs (Mean $\pm$ SD over 15 runs)}
    \label{tab:vulnerability_discovery}
    \scriptsize
    \begin{tabular*}{\columnwidth}{@{\extracolsep{\fill}}lcccc@{}}
        \toprule
        \textbf{DUT (Model)} & \textbf{Baseline Acc.} & \textbf{Critical Bits} & \textbf{Final Acc.} & \textbf{Time (hrs)} \\
        \midrule
        GPT-2 Large & 30.5\% & 5.1 $\pm$ 0.6 & 0.34\% & 89 $\pm$ 8 \\
        LLaMA 3.1 8B & 69.9\% & 5.3 $\pm$ 0.7 & 0.18\% & 312 $\pm$ 18 \\
        DeepSeek-V2 7B & 71.3\% & 5.8 $\pm$ 1.1 & 0.22\% & 156 $\pm$ 12 \\
        \midrule
        \textbf{Average} & -- & \textbf{5.4 $\pm$ 0.8} & \textbf{0.25\%} & \textbf{186 $\pm$ 13} \\
        \bottomrule
    \end{tabular*}
    \vspace{-2mm}
\end{table}

\subsection{RQ2: Reliability-Aware Design Insights}
\label{subsec:vulnerability_analysis}

Prior work has shown that state-of-the-art AI accelerators can exhibit sparse and highly localized vulnerability patterns; finding these weaknesses in the early design stages is critical for enabling efficient, targeted protection strategies~\cite{nazari2024forget, reagen2018ares, santos2019analyzing}. Motivated by this, this investigation demonstrates RIFT's capability to provide the actionable, quantitative data necessary for reliability-aware hardware Design Space Exploration (DSE).
In Table~\ref{tab:vulnerability_discovery}, we present the results of using RIFT to evaluate the presence of sparse vulnerabilities in our DUTs, quantifying the minimal fault set size required to induce catastrophic functional failure. The results demonstrate that, for all evaluated models, a complete collapse in functional accuracy (greater than 99\% degradation) can be induced by perturbing an average of only \textbf{5.4 $\pm$ 0.8 critical bits}. This finding substantiates the existence of sparse and high-impact failure modes and underscores the need for fault assessment methodologies that prioritize the identification of such critical fault combinations over analyses of diffuse and random faults.
Prior studies have identified attention mechanisms and normalization layers as particularly vulnerable components in AI accelerators~\cite{reagen2018ares, li2017understanding}. Our systematic analysis provides precise quantification of this phenomenon, finding that 88.5\% of critical faults concentrate in attention mechanisms (47.3\%) and normalization layers (41.2\%), while feed-forward networks remain comparatively robust—highlighting clear fault-sensitive hotspots for targeted protection strategies.

In Table~\ref{tab:protection_analysis}, we quantify the cost-effectiveness of different protection schemes against the worst-case faults identified by RIFT. While uniform TMR provides the highest coverage (99.2\%), its prohibitive 205\% area overhead results in a low cost-effectiveness score of 0.5. By contrast, a RIFT-guided selective ECC strategy—targeting only the vulnerable regions identified in our analysis—achieves 88.5\% fault coverage with just 13.8\% overhead, yielding a cost-effectiveness of 6.4, which is 12.8$\times$ higher than TMR. These results highlight the engineering advantage of data-driven, targeted protection over uniform fault coverage.
\begin{table}[ht]
    \centering
    \caption{DSE
    for Protection Strategies. 
    AO = Area Overhead, FC = Fault Coverage, CE = Cost-Effectiveness (Coverage/Area)}
    \label{tab:protection_analysis}
    \scriptsize
    \begin{tabular*}{\columnwidth}{@{\extracolsep{\fill}}lcccc@{}}
        \toprule
        \textbf{Strategy} & \textbf{AO} & \textbf{FC} & \textbf{CE} & \textbf{Notes} \\
        & \textbf{(\%)} & \textbf{(\%)} & \textbf{(Cov/Area)} & \\
        \midrule
        No Protection & 0 & 0 & -- & Baseline \\
        Parity (Uniform) & 6.3 & 0\tablefootnote{Parity: detection only, 0\% functional coverage. Overhead from synthesis studies~\cite{micron2023ecc}.} & 0.0 & Detect Only \\
        ECC SECDED (Uniform) & 18.7 & 95.1 & 5.1 & Std Correction \\
        ECC ChipKill (Uniform) & 31.4 & 98.7 & 3.1 & Adv Correction \\
        TMR (Uniform) & 205.0 & 99.2 & 0.5 & Max Redundancy \\
        \midrule
        \textbf{RIFT-Guided ECC} & \textbf{13.8} & \textbf{88.5} & \textbf{6.4} & \textbf{Targeted} \\
        \bottomrule
    \end{tabular*}
    \vspace{-2mm}
\end{table}

\begin{table}[t]
    \centering
    \caption{Statistical Robustness of RIFT's Key Performance Metrics (N=15 independent runs)}
    \label{tab:statistical_robustness}
    \scriptsize
    \begin{tabular}{lcccc}
        \toprule
        \textbf{Metric} & \textbf{Mean $\pm$ SD} & \textbf{95\% CI} & \textbf{p-value}\tablefootnote{Wilcoxon signed-rank test vs. best baseline (GenBFA).} & \textbf{Cohen's d} \\
        \midrule
        Critical Bits & 5.4 $\pm$ 0.8 & [5.0, 5.8] & $< 0.001$ & 2.73 \\
        Fault Coverage (\%) & 91.7 $\pm$ 2.1 & [90.6, 92.8] & $< 0.001$ & 3.85 \\
        Efficiency (Cov/hr) & 0.490 $\pm$ 0.034 & [0.472, 0.508] & $< 0.001$ & 4.12 \\
        \bottomrule
    \end{tabular}
    \vspace{0.1cm}
    \vspace{-4mm}
\end{table}

\subsection{RQ3: Framework Consistency and Scalability}
\label{subsec:consistency_scalability}

A consistent fault assessment framework must reliably identify vulnerabilities of similar magnitude and impact across independent trials, even with different random initializations. To validate RIFT's consistency, we conducted 15 independent runs of the entire discovery process for each DUT. The statistical analysis of these runs is detailed in Table~\ref{tab:statistical_robustness}. The results show remarkable consistency in the framework's discovery capability. For instance, the number of critical bits discovered is tightly clustered around a mean of 5.4 $\pm$ 0.8, with a narrow 95\% confidence interval of [5.0, 5.8]. Similarly, the final fault coverage achieved by the framework is highly stable at \textbf{91.7\% $\pm$ 2.1\%}. The highly significant p-values ($<$ 0.001) and large effect sizes (Cohen's $d$ $>$ 2.5) further confirm that RIFT consistently and reliably identifies critical failure modes.

\begin{figure}[b]
    \centering
    \includegraphics[width=0.8\linewidth]{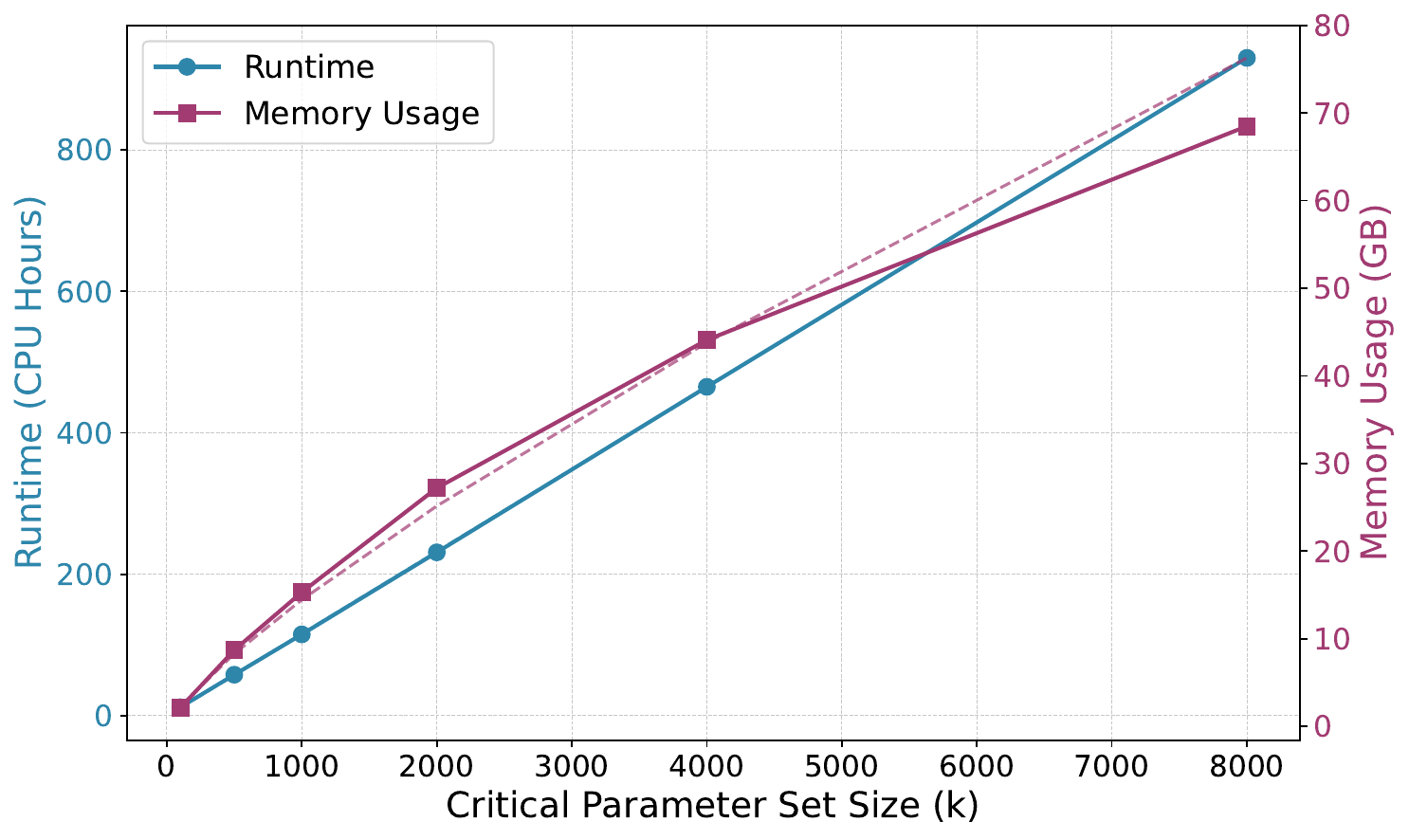}
    \caption{Scalability of RIFT: runtime grows linearly ($R^2=0.998$), while memory usage scales super-linearly ($O(k^{1.3})$, $R^2=0.995$)
    }
    \label{fig:scalability_results}
    \vspace{-4mm}
\end{figure}

Beyond consistency, a practical fault assessment framework must be scalable, meaning its computational resource requirements should grow predictably with the complexity of the DUT. We analyzed RIFT's computational scaling characteristics by evaluating its performance on models of varying sizes, with the results presented in Figure~\ref{fig:scalability_results}. The analysis confirms that the framework's runtime scales with a strong and predictable \textbf{linear relationship} to the number of parameters in the target fault-sensitive hotspot $k$, as defined in Phase 2 of our methodology. This is evidenced by a near-perfect linear fit (R² $>$ 0.99). The framework's memory requirements grow at a modest, \textbf{super-linear rate}, empirically determined to be approximately $O(k^{1.3})$. To provide a concrete example, for the largest tested configuration where a critical set size of $k=8000$ parameters was selected, the framework required approximately 930 CPU hours and 68.5 GB of memory. This predictable, non-exponential scaling for both time and memory ensures that RIFT remains a viable and practical tool for the fault assessment of even larger, next-generation AI accelerators.

\section{Framework Analysis and Ablation Studies}
\label{sec:ablation_studies}

In this section, we present comprehensive ablation studies validating RIFT’s design choices and parameters sensitivity, demonstrating that each component enhances fault assessment performance and that the framework remains robust across diverse parameter settings. Due to the high computational cost, these detailed studies were conducted primarily on the LLaMA 3.1 8B model, with key findings subsequently validated across the other architectures.

\subsection{Analysis of the Hybrid Sensitivity Metric}

\begin{figure}[t]
    \centering
    \includegraphics[width=0.8\linewidth]{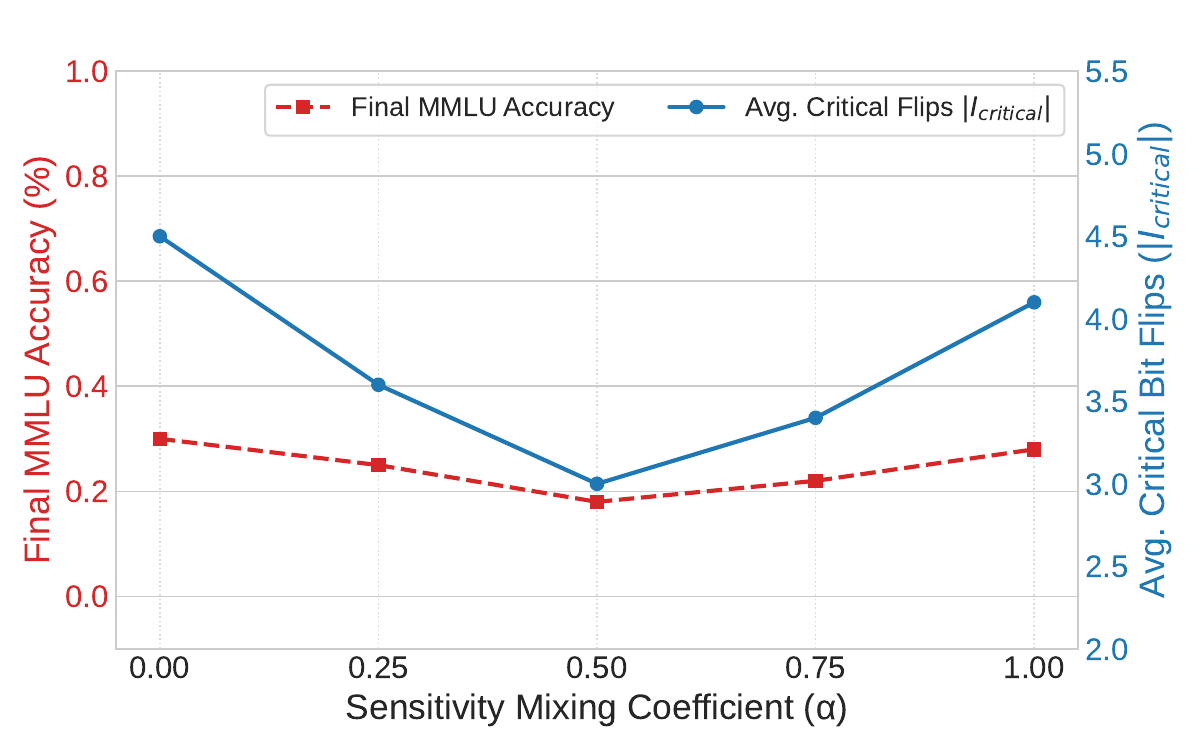}
    \caption{Impact of 
    $\alpha$ on fault assessment effectiveness.
    }
    \label{fig:ablation_alpha}
    \vspace{-4mm}
\end{figure}

We evaluated the effect of the sensitivity mixing coefficient $\alpha$ from Equation~\ref{eq:sensitivity}, which balances static magnitude and dynamic gradient information. Varying $\alpha$ from 0.0 (magnitude-based) to 1.0 (gradient-based) while keeping other parameters fixed, we measured the minimal critical fault set required to trigger catastrophic functional failure. As shown in Figure~\ref{fig:ablation_alpha}, the pure magnitude-based setting ($\alpha=0.0$) required 7.1 $\pm$ 0.8 faults, whereas the gradient-based setting ($\alpha=1.0$) reduced this to 6.4 $\pm$ 0.6. The hybrid configuration ($\alpha=0.5$) achieved the greatest efficiency, identifying only \textbf{5.0 $\pm$ 0.4 faults}, a \textbf{29\% reduction} relative to the magnitude-based baseline. Moreover, $\alpha$ values within [0.25, 0.75] consistently outperformed the pure approaches, indicating robustness to parameter variation.
These results demonstrate that the hybrid sensitivity metric effectively integrates static and dynamic information, yielding a superior guiding signal for fault discovery. 

\begin{table}[t]
    \centering
    \vspace{-2mm}
    \caption{Ablation Analysis of RIFT
    on LLaMA 3.1 8B.}
    \label{tab:architecture_ablation}
    \scriptsize
    \begin{tabular*}{\columnwidth}{@{\extracolsep{\fill}}lcc@{}}
        \toprule
        \textbf{Configuration} & \textbf{Critical Faults Found} & \textbf{Episodes to Converge} \\
        \midrule
        Complete RIFT & 5.0 $\pm$ 0.4 & 50 $\pm$ 3 \\
        RL-Only (Same Budget) & 47.3 $\pm$ 8.2 & 50 (Fixed) \\
        RL-Only (To Converge) & 5.2 $\pm$ 0.6 & 890 $\pm$ 67 \\
        \bottomrule
    \end{tabular*}
    \vspace{-3mm}
\end{table}

\subsection{Validation of the Three-Phase Architecture}

We evaluated RIFT’s three-phase design space reduction against an \emph{RL-Only} baseline, where the agent is initialized with random parameters, bypassing Phases 1 and 2. Table~\ref{tab:architecture_ablation} reports the critical fault set size and computational effort required for convergence. Under the same 50-episode budget, RL-Only yields an ineffective set averaging 47.3 faults, whereas RIFT identifies a minimal set of 5.2 faults. Achieving comparable efficiency with RL-Only requires 890 episodes—over \textbf{17$\times$ more effort}.
These results confirm that vulnerability profiling is not a mere preprocessing step but a critical algorithmic component. Sensitivity-guided initialization provides essential prior knowledge, enabling efficient exploration of the high-dimensional fault space and ensuring scalable fault assessment.

\subsection{RL Parameter Sensitivity}

\begin{figure}[t]
    \centering
    \includegraphics[width=0.8\linewidth]{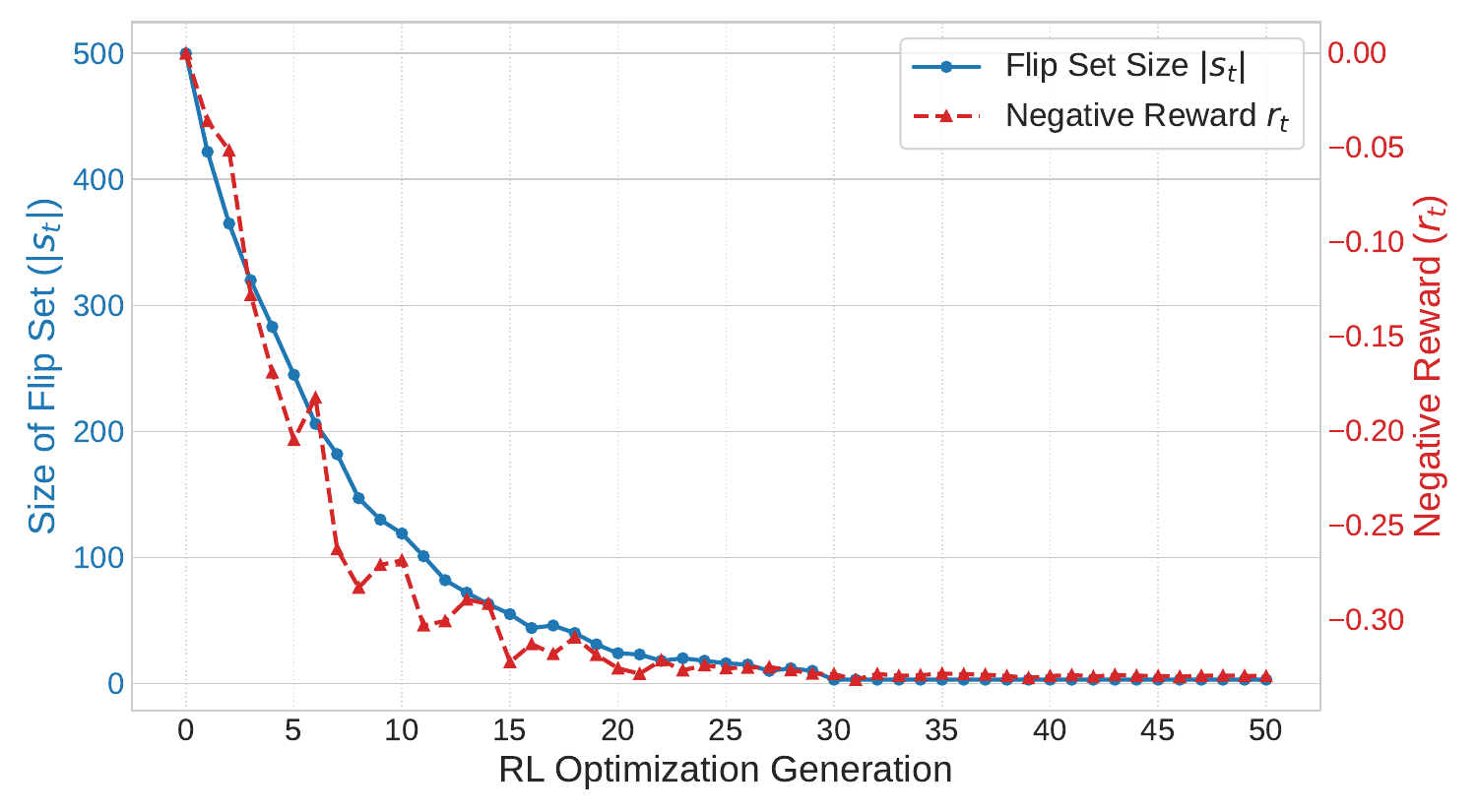}
    \caption{RIFT RL parameter sensitivity analysis
    }
    \label{fig:rl_sensitivity}
    \vspace{-4mm}
\end{figure}

In this subsection, we analyze RIFT’s sensitivity to its core RL hyperparameters—the number of training episodes ($G$) and the exploration rate ($\epsilon$). Figure~\ref{fig:rl_sensitivity} shows their effect on the final critical fault set size. Performance saturates around $G=50$, with further increases providing diminishing returns, while reducing $G$ significantly degrades results. For $\epsilon$, RIFT remains stable across values in [0.05, 0.30], with under 5\% variation in fault set size.
These findings confirm that RIFT’s efficiency does not rely on fragile hyperparameter tuning but stems from its intrinsic design. Such robustness enhances its practicality, enabling deployment across diverse fault assessment scenarios without extensive case-specific optimization.

\vspace{-1mm}
\subsection{Generalization Across Architectures}

To assess generality, we repeated the ablation studies across three DUTs, with results summarized in Table~\ref{tab:generalization_ablation}. The hybrid sensitivity metric ($\alpha=0.5$) consistently outperformed pure magnitude- and gradient-based approaches, improving fault set minimality by 24–31\%. Similarly, three-phase architecture outperformed RL-only baseline by 3.8–4.6$\times$ across all models.
These findings confirm that RIFT’s advantages are consistent across architectures, demonstrating that its superior efficiency stems from principled algorithmic design rather than model-specific artifacts. This establishes RIFT as a robust, general-purpose fault assessment framework for diverse AI hardware.

\begin{table}[htbp]
    \centering
    \caption{Cross-Architecture Validation of Key Ablation Findings.}
    \label{tab:generalization_ablation}
    \scriptsize
    \begin{tabular}{lcc}
        \toprule
        \textbf{DUT (Model)} & \textbf{Benefit of Hybrid Metric}\tablefootnote{Reduction in critical fault count vs. best pure approach ($\alpha=0$ or $1$).} & \textbf{Benefit of 3-Phase Arch.}\tablefootnote{Efficiency improvement (Cov/Time) vs. RL-only baseline.} \\
        \midrule
        GPT-2 Large & 24\% & 3.8× \\
        LLaMA 3.1 8B & 29\% & 4.2× \\
        DeepSeek-V2 7B & 31\% & 4.6× \\
        \bottomrule
    \end{tabular}
    \vspace{0.1cm}
    \vspace{-5mm}
\end{table}

\section{Conclusion}
\label{sec:conclusion}

This paper introduced \textbf{RIFT}, a novel reinforcement learning framework that directly addresses the critical fault assessment scalability crisis in modern AI accelerator design. By reformulating the intractable search for worst-case faults as a sequential decision-making problem, RIFT provides an automated, intelligent methodology that overcomes the cost and coverage limitations of traditional fault assessment techniques.
Our comprehensive evaluation, validated on industrial-grade AI workloads using NVIDIA A100 GPUs, demonstrates RIFT's significant contributions to Design Automation. The framework accelerates fault assessment workflows, achieving a \textbf{2.2$\times$} speedup in efficiency over state-of-the-art evolutionary methods and reducing the required test vector volume by over \textbf{99\%} compared to widely adopted random fault injection. Beyond efficiency, RIFT enables true reliability-aware design space exploration, providing the empirical data to improve hardware cost-effectiveness by over \textbf{12.8$\times$} through targeted, selective protection strategies.

This work establishes RIFT as a robust fault assessment methodology, while also opening several directions for future research. Beyond LLMs, applying RIFT to other large-scale architectures, such as vision transformers and diffusion models, will further validate its generality. Extending the framework to capture more complex fault models, including transient logic errors and timing-related faults, is another natural progression.
Looking ahead, our long-term vision is to integrate RIFT’s intelligent exploration engine into fully automated, reliability-aware synthesis flows, where quantitative fault analysis directly guides physical design optimization. Ultimately, RIFT provides a principled foundation for automating the design and fault assessment of reliable AI hardware—an essential capability for advancing safety-critical semiconductor applications.

\section{ACKNOWLEDGMENTS}
This material is based upon work supported by the National Science Foundation (NSF) under Award Numbers: CCF-2323819. Any opinions, findings, conclusions, or recommendations expressed in this publication are those of the authors and do not necessarily reflect the views of the NSF.

\balance 
\bibliographystyle{IEEEtran}
\bibliography{bib/date, bib/dac, bib/ref}

\end{document}